\def\year{2018}\relax
\documentclass[letterpaper]{article} 
\usepackage{aaai18}  
\usepackage{times}  
\usepackage{helvet}  
\usepackage{courier}  
\usepackage{url}  
\usepackage{graphicx}  
\usepackage[vlined,ruled,linesnumbered]{algorithm2e}
\usepackage{amsmath}
\usepackage{mathrsfs}
\usepackage{array,booktabs}
\usepackage[caption=false]{subfig}

\begin{document}
%
\title{Motif-based Rule Discovery for  Predicting \\Real-valued Time Series\thanks{The work remains incomplete and will be further refined. Read it on your own risk :).}\thanks{We are grateful to Leye Wang, Junyi Ma, Zhu Jin, and Siqi Yang for their invaluable help.}\thanks{We are also grateful for the reviewers in AAAI18.}
}

\author{Yuanduo He, Xu Chu, Juguang Peng, Jingyue Gao, Yasha Wang\\
	 Key Laboratory of High Confidence Software Technologies,\\
	Ministry of Education, Beijing 100871, China\\
	\{ydhe, chu\_xu, pgj.pku12, gaojingyue1997, yasha.wang\}@pku.edu.cn
}
\nocopyright
\maketitle
\begin{abstract}
Time series prediction is of great significance in many applications and has attracted extensive attention from the data mining community. Existing work suggests that for many problems, the shape in the current time series may correlate an upcoming shape in the same or another series. Therefore, it is a promising strategy to associate two recurring patterns as a \textit{rule}'s antecedent and consequent: the occurrence of the antecedent can foretell the occurrence of the consequent, and the learned shape of consequent will give accurate predictions. Earlier work employs symbolization methods, but the symbolized representation maintains too little information of the original series to mine valid rules. The state-of-the-art work, though directly manipulating the series, fails to segment the series precisely for seeking antecedents/consequents, resulting in inaccurate rules in common scenarios. In this paper, we propose a novel motif-based rule discovery method, which utilizes motif discovery to accurately extract frequently occurring consecutive subsequences, i.e. motifs, as antecedents/consequents. It then investigates the underlying relationships between motifs by matching motifs as rule candidates and ranking them based on the similarities. Experimental results on real open datasets show that the proposed approach outperforms the baseline method by 23.9\%. Furthermore, it extends the applicability from single time series to multiple ones.
\end{abstract}

\section{Introduction}

The prediction of real-valued time series is a topic of great significance and colossal enthusiasm in the data mining community, and has been applied extensively in many research areas \cite{hamilton1994time,esling2012time}. Most of the work in literature aims at modeling the dependencies among variables, and forecasting the next few values of a series based on the current values \cite{makridakis2008forecasting}. However, other work suggests that for many problems it is the shape of the current pattern rather than actual values that makes the prediction \cite{das1998rule}. For clarity, we call the latter one, forecasting by shape, \textit{rule-based prediction}, which is the subject of this paper.

Informally\footnote{A formal definition is given in the Preliminaries section.}, a rule $A\Rightarrow_\tau B$ associates a pattern $A$ as the antecedent with a pattern $B$ as the consequent in an upper-bounding time interval $\tau$. The rule-based prediction works as follows: if a shape resembling $A$ is observed in a series, then a shape resembling $B$ is supposed to appear in the same or another series, within the time interval $\tau$. Usually, a rule refers to an ``authentic  rule'', an underlying relationship between two events implied by the two shapes. 


In most work, firstly the subsequence clustering method is employed to symbolize/discretize the series, and then the symbolic series rule discovery method is applied to find rules from real-valued series \cite{das1998rule}. However, such work has met limited success and ends up discovering spurious rules (e.g. ``good'' rules discovered from random walk data), because the symbol representation by common symbolization methods \textit{is independent with} the raw real-valued series \cite{keogh2005clustering}. In conclusion, the symbolized time series maintains little information about the original series to mine valid rules. 

The state-of-the-art work \cite{shokoohi2015discovery} directly manipulates the series under the assumption that \textit{a rule is contained in a subsequence}. It first selects a subsequence and then splits it into a rule's antecedent and consequent. However, usually there is an interval between a rule's antecedent and consequent. The splitting method will append the extra interval series to the antecedent/consequent, which fails to segment precisely for antecedents/consequents and results in rules with a lower prediction performance. Furthermore, it cannot be applied to find rules from multiple series, i.e. a shape in one series predicting a shape in another series.

Only when the observed object presents repeatability, can predictions be valid/reasonable.
We believe that a serviceable rule should appear repeatedly in a real-valued time series and therefore its antecedent and consequent must be recurring patterns of the series. 
Recent studies \cite{vahdatpour2009toward,brown2013dictionary,mueen2014time} show that Motifs, \textit{frequently occurring patterns in time series} \cite{patel2002mining}, contain key information about the series. Therefore, we could utilize motif discovery methods to find all distinct motifs which accurately represent antecedent and consequent candidates of rules.

\begin{figure}[bt]
	\centering
	\includegraphics[width=1\linewidth]{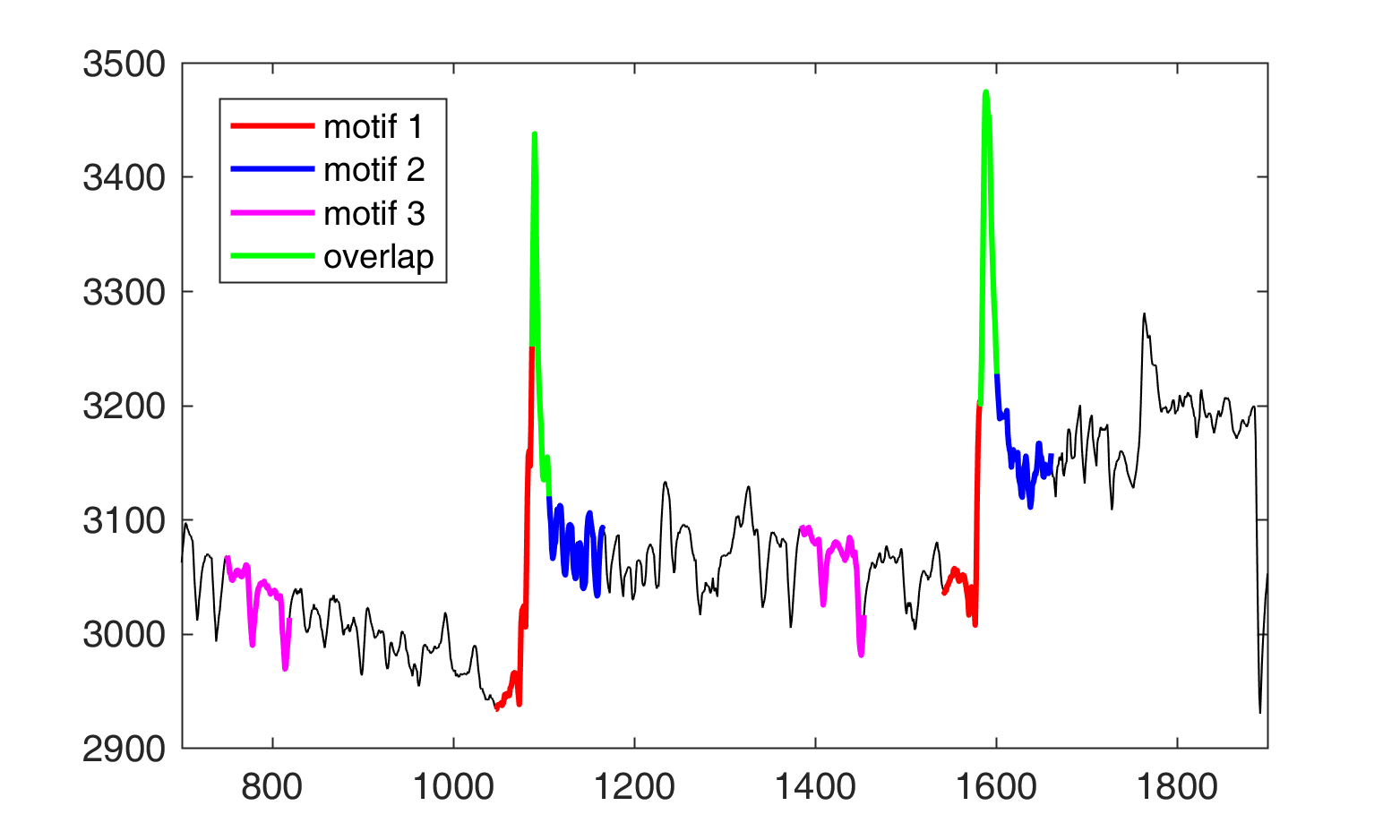}
	\caption{The electrical penetration graphs (EPG) data monitoring of the probing behavior of leafhoppers. There are three motifs in the series, motif 1 and motif 2 are overlapped in the green curves.}
	\label{fig:motifs}
\end{figure}

\begin{figure}[bt]
	\centering
	\includegraphics[width=0.9\linewidth]{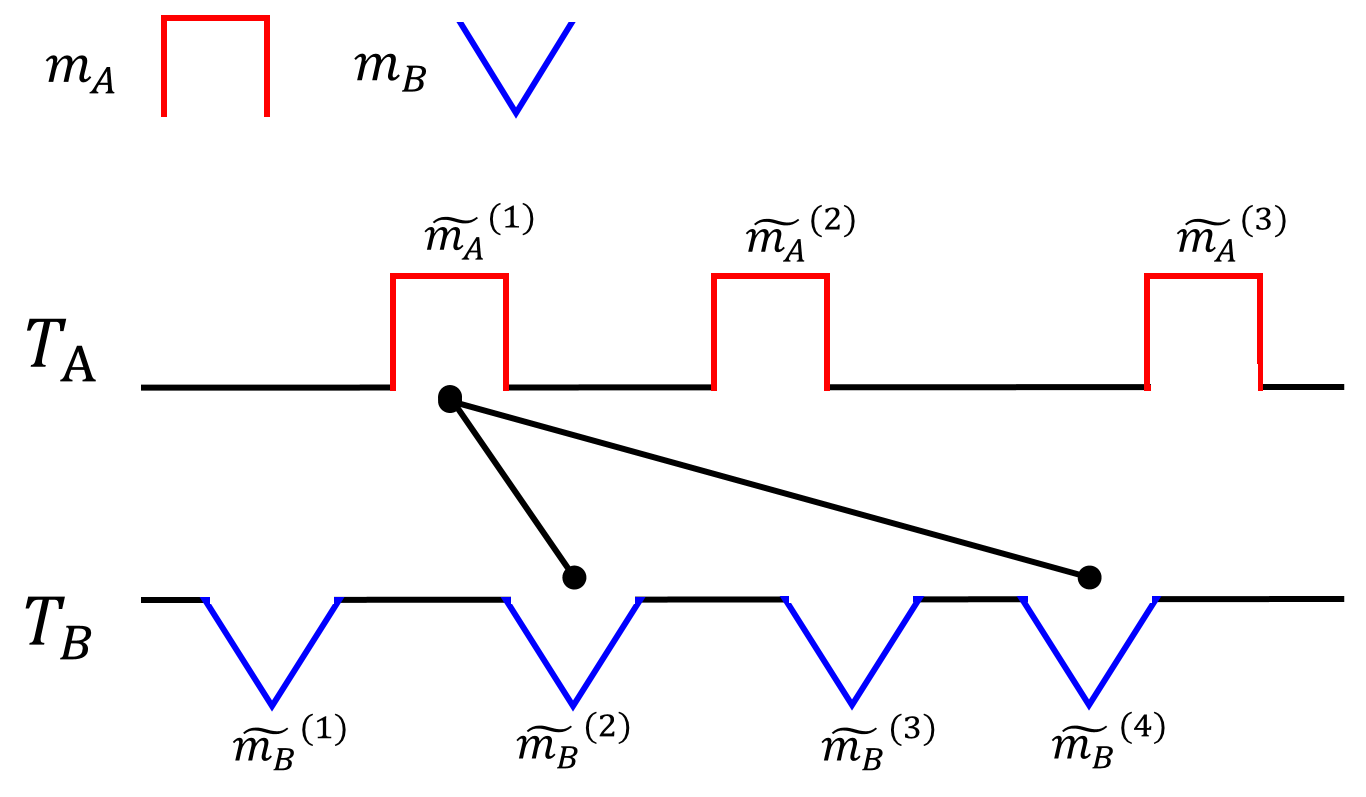}
	\caption{An underlying rule with respect to motif $m_A$ and $m_B$. $T_A$ and $T_B$ are two series. $\tilde{m_A}^{(i)}$ and $\tilde{m_B}^{(j)}$ are \textit{motif instances} of $m_A$ and $m_B$ in $T_A$ and $T_B$, respectively. $\tilde{m_A}^{(1)}$ with $\tilde{m_B}^{(2)}$ or $\tilde{m_B}^{(4)}$  can form different \textit{rule instances}, i.e. the two solid lines.}
	\label{fig:example}
\end{figure}

However, three challenges are confronted: (1) there could be quite a few (overlapping) motifs in a single series (see Figure \ref{fig:motifs}), let alone the dataset containing multiple series. How to screen for effective motifs as antecedents and consequents? (2) there could be many instances for two given effective motifs and different combinations of them will lead to different instances of a rule (see Figure \ref{fig:example}). How to identify the best combinations for the underlying rule? (3) How to rank different rules given the instances of each rule?

In this paper, we present a rule-based prediction method for real-valued time series from the perspective of motifs. For each pair of motifs as a candidate rule, a heuristic-matching-based scoring algorithm is developed to investigate the connection between them. 



We summarize contributions of this paper as follows:
\begin{itemize}
	\item We propose a novel rule discovery approach from the perspective of motifs. It can find rules with higher prediction performance and can also be applied to multiple time series.
	\item We develop a novel heuristic matching algorithm in search of the best combinations of motif instances. To accommodate the heuristic matching result, we modify Yekta's scoring method leveraging Minimum Description Length (MDL) principle to evaluate each rule \cite{shokoohi2015discovery}.
	\item We evaluate our work on real open datasets, and the experimental results show that our method outperforms state-of-the-art work by $23.9\%$ on average\footnote{This experiment is performed on a single dataset, which is inadequate. We will refine it. Please see future work.}. Moreover, rules in multiple series can also be discovered. 
\end{itemize}

\section{Related Work}
Early work extracts association rules from frequent patterns in the transactional database \cite{han2011data}. Das et al. are the first to study the problem of finding rules from real-valued time series in a symbolization-based perspective \cite{das1998rule}, followed by a series of work \cite{sang2001discovering,wu2004online}. 

However, it has been widely accepted that few work dealing with real-valued series discovers valid rules for prediction. Because the symbolized series by common symbolization methods, including subsequence clustering and Piecewise Linear Approximation (PLA), cannot effectively represent the original series. \cite{struzik2003time,keogh2005clustering,shokoohi2015discovery}. Eamonn Keogh et al. (2005) demonstrate that clustering of time series subsequences is meaningless, resulting from the independency between the clustering centers and the input. Shokoohi-Yekta et al. (2015) point out that two time series can be differ only by a small discrepancy, yet have completely different PLA representations. Therefore, the symbolized series is irrelevant with the original real-valued one and the failure of finding valid rules is doomed.

The state-of-the-art work (Y15) directly manipulates the real-valued series \cite{shokoohi2015discovery}. Their method is found on the assumption that a rule is contained in a subsequence, which splits a subsequence into the rule's antecedent and consequent. Usually, there is an interval between a rule's antecedent/consequent, and the splitting method will append the extra series to the antecedent/consequent. The complicated diversity of the intervals could result in rules with bad prediction performance. Besides, the splitting method cannot be applied to discover rules from two series either, since a motif is a subsequence in a single series. 

The difference between Y15 with our work  can be seen from the perspective of the different usages of motifs. In Y15, it is noted that ``\textit{In principle we could use a brute force search, testing all subsequences of T}'', which means that it can work without motifs; but our method cannot work without motifs, since we are trying to relate pairs of them. Even if Y15 method uses motifs to ``\textit{provide a tractable search}'', it is still different from ours. As it mentioned, ``\textit{a good rule candidate must be a time series motif in T}'', Y15 treats motifs as rule candidates, while our method takes motifs as candidates of antecedents and consequents, as we noted that ``\textit{we could utilize motif discovery methods to find all distinct motifs which accurately represent antecedent and consequent candidates of rules}''.

\section{Preliminaries}
In this paper, we consider the problem of rule discovery from real-valued time series for rule-based prediction. Without loss of generality, the rule discovery problem is exemplified using two time series $T_A$ and $T_B$, since finding rules from over two time series can be treated pairwisely. It can also be applied to the situation of only one series by letting $T_A = T_B$. We begin with the definition of a rule and its instance.

\noindent \textbf{Definition 1 (Rule).} \textit{A rule $r$ is a 4-tuple $(m_A, m_B, \tau, \theta)$. $m_A, m_B$ are  the subsequences of two real-valued time series $T_A, T_B$, respectively, and also the antecedent and consequent of the rule $r$. $\tau$ is a non-negative value\footnote{In real applications, $\tau$ is  a predefined value according to the series, which avoids nonsense rules with infinity max time interval.} indicating the max length of time interval between the $m_A$ and $m_B$. $\theta$ is a non-negative value as the trigger threshold for firing the rule.}

\noindent \textbf{Definition 2 (Instance).} \textit{A rule $r$'s instance $e$ is a 3-tuple $(\tilde{m_A}, \tilde{m_B}, \tilde{\tau})$. $\tilde{m_A}$ is the $m_A$-like subsequence observed in series $T_A$ subject to $d(\tilde{m_A}, m_A) < \theta$. $\tilde{m_B}$ is the $m_B$-like subsequence observed later than $\tilde{m_A}$ in series $T_B$. $\tilde{\tau}$ is the time interval between $\tilde{m_A}$ and $\tilde{m_B}$.}

By the definitions, given a rule $r = (m_A, m_B, \tau, \theta)$, if a subsequence $\tilde{m_A}$ of $T_A$ is observed and $d(\tilde{m_A}, m_A)<\theta$, the rule $r$ is fired and a subsequence of $\tilde{m_B}$ similar with $m_B$ is supposed to be observed from $T_B$ within the subsequent $\tau$-time interval\footnote{In fact, $\tilde{m_B}$ is observed after $\tilde{\tau}$ time.}. Notice that there are no constraints for $\tilde{\tau}$ and $d(\tilde{m_B}, m_B)$ in Definition 2, since an instance with $\tilde{\tau} > \tau$ or $d(\tilde{m_B}, m_B) \gg 0$ can be viewed as a ``bad'' instance for $r$. An example is shown in Figure \ref{fig:example}.

Intuitively, if a rule is good, it must have many supporting instances in the time series, and then $\tilde{m_A}$ and $\tilde{m_B}$ will be \textit{frequently occurring patterns}. By the definition of the motif \cite{patel2002mining}, they are actually the motifs of $T_A$ and $T_B$, respectively. Inspired by that, we make the following assumption.

\noindent \textbf{Assumption 1.} \textit{If $r = (m_A, m_B, \tau, \theta)$ is a rule, then $m_A \in M_A$, $m_B \in M_B$, where $M_A$ and $M_B$ are motif sets of $T_A$ and $T_B$.}

Based on the assumption, the rule discovery problem can be formulated as follows. \textit{Given two time series $T_A, T_B$ with their motif sets $M_A$ and $M_B$, find top-K rules $r = (m_a, m_b, \tau, \theta)$, where $m_a \in M_A$, and $m_b \in M_B$.} Efficient algorithms for motif discovery have already been developed, and we could choose the widely-applied MK algorithm \cite{mueen2009exact}\footnote{For concise, the footnotes of all Ks are omitted throughout this paper. In fact, Ks can be different.}.

Directly, we present the top-level algorithm \ref{alg:top}. It first finds the motif set for each series in line 1 by  MK algorithm. Line 2 sorts and returns the top K motifs according to the domain knowledge of the series. Then it traverses every pair of motifs, and scores the corresponding rule in line 3 to 4. It finally returns the best K rules. 


\begin{algorithm}[t]
\DontPrintSemicolon
\KwIn{$T_A$ and $T_B$ are two time series.}
\KwOut{$Res$ is K best rules.}
\BlankLine
$M_A, M_B \leftarrow \textit{motifs}(T_A), \textit{motifs}(T_B)$\;
$M_A, M_B \leftarrow \textit{sortK}(M_A), \textit{sortK}(M_B)$\;
\ForAll{$(m_A, m_B) \in M_A \times M_B$}{
	\textit{score}($m_A$, $m_B$, $T_A$, $T_B$)\;
}
$Res \leftarrow$ \textit{topK\_score\_rules}\;

\caption{Find top-K rules.}
\label{alg:top}
\end{algorithm}

\section{Methodologies}
Given a pair of motif $m_a$ and $m_b$, the score algorithm aims at evaluating a rule candidate $r$ based on the instances in the training data. We will use the example in Figure \ref{fig:example} to illustrate. The scoring approach consists of three steps:

\begin{enumerate}
	\item \textit{Find out all $m_A, m_B$-like patterns in $T_A, T_B$.} In this step, we use the sliding window method to select similar patterns by setting a threshold. In Figure \ref{fig:example}, three $m_A$-like patterns $\tilde{m_A}^{(i)} (i=1,2,3)$ and four $m_B$-like patterns $\tilde{m_B}^{(i)} (i=1,..,4)$ are discovered;
	\item \textit{Match $m_A,m_B$-like patterns into rule instances, and search for the matching result that can support the rule most.} A brutal search is not only intractable but also lack of robustness. Instead, we propose a heuristic matching algorithm according to the belief that a rule is preferred when it has (1) many instances and (2) a short max-length time interval. 
	The lines (both dotted and solid) in Figure \ref{fig:graph} are all possible instances and the solid ones $\{e_{1, 2}, e_{2, 3}\}$ are the best instances, because it has the most number of instances (i.e. 2), and the average length of time intervals is also the smallest.
	\item \textit{Score each instance with respect to the rule $r$, and then integrate to the final score.} In this step, instead of Euclidean distance, we follow Yekta's scoring method and further consider the ratio of antecedents being matched, i.e. $2/3$. In the example, the two best instances $\{e_{1, 2}, e_{2, 3}\}$ are evaluated respectively based on MDL principle. The final score for $r$ is the sum of  $\{e_{1, 2}, e_{2, 3}\}$'s results multiplied by $2/3$. 
\end{enumerate}

\subsection{Step 1. Motif-like Pattern Discovery}




MK algorithm \cite{mueen2009exact} returns pairs of subsequences which are very similar with each other as motifs. In this step, given a motif $m$ (a pair of subsequences) of a time series $T$, we need to search for all $m$-like non-overlapped subsequences in $T$. 

A direct approach based on sliding window method is as follows: (1) search all subsequences of the same length with $m$ and calculate the distance between them and $m$; (2) set an appropriate threshold $\theta'$ to filter the subsequences with distance smaller than $\theta'$; (3) sort the remaining subsequences by distance and remove the overlap ones with larger distances. After that, the motif-like patterns are chosen as the non-overlapped subsequences with small distance. 

The rule threshold $\theta$ is set as the threshold for selecting antecedent-like subsequences. The complexity of sliding window method is determined by the number of windows and the distance computation procedure. In this step, the complexity is $O(|m|\cdot|T|)$.

\subsection{Step 2. Heuristic Matching}
The patterns found in step 1 can be combined into pairs as instances of a rule. This step aims at finding the best instance set that support the rule, which should satisfy the following conditions: (1) its cardinality is the largest among all possible sets; (2) the average length of interjacent intervals of instances is the smallest. The two condition come from the belief about what a good rule is. 

\textbf{Modeling.} To formulate the problem concretely, we introduce the following notations and construct a weighted bipartite graph.

$\widetilde{M_A} = \{\tilde{m_A}^{(1)}, ..., \tilde{m_A}^{(p)}\}, \widetilde{M_B} = \{\tilde{m_B}^{(1)}, ..., \tilde{m_B}^{(q)}\}$ are the sets containing all subsequences similar with the rule's antecedent $m_A$ and consequent $m_B$, respectively.   
 
$E = \{e_{i, j} = (\tilde{m_A}^{(i)}, \tilde{m_B}^{(j)}) | 1 \le i \le p, 1 \le j \le q,  0 < t(\tilde{m_B}^{(j)}) - t(\tilde{m_A}^{(i)}) < \tau\}$, where function $t(\cdot)$ returns the occurrence time of the pattern. $E$ is the set of all \textit{feasible} instances, since the antecedent must appear before the consequent and the interval between them cannot be too large. It imposes a structure on the set $E$ that given $\tilde{m_A}^{(i)}$, for $\forall j$ such that $\tau > t(\tilde{m_B}^{(j)}) - t(\tilde{m_A}^{(i)}) > 0$, then $e_{i, j} \in E$. Besides, let $w_{i, j} = t(\tilde{m_B}^{(j)}) - t(\tilde{m_A}^{(i)})$ measure the length of interjacent interval of the instance $e_{i, j}$. 

\begin{figure}[t]
	\centering
	\includegraphics[width=0.7\linewidth]{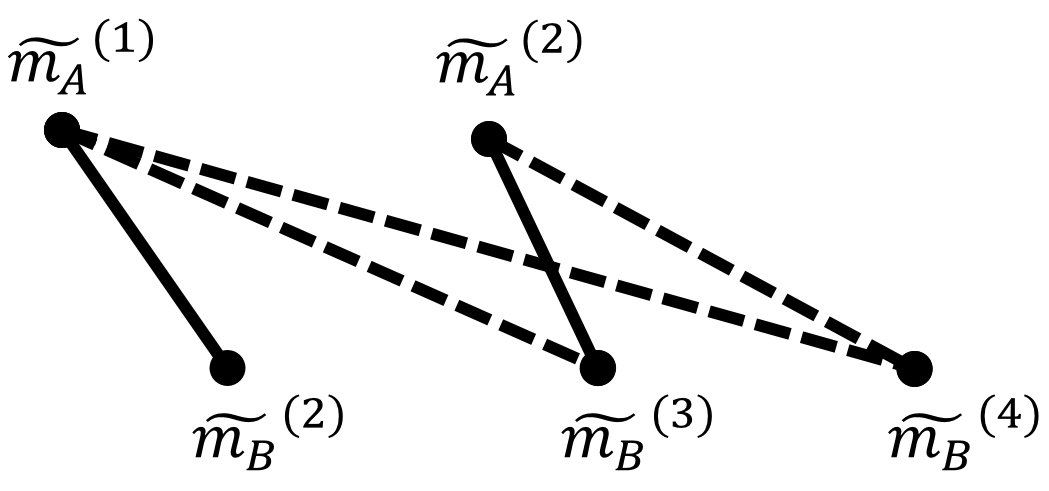}
	\caption{The graph $G$ for the example in Figure \ref{fig:example}. $\widetilde{M_A} = \{\tilde{m_A}^{(1)}, \tilde{m_A}^{(2)}\}$, $\widetilde{M_B} = \{\tilde{m_B}^{(2)}, \tilde{m_B}^{(3)}, \tilde{m_B}^{(4)}\}$, and $E=\{e_{1,2},e_{1,3},e_{1,4},e_{2,3},e_{2,4}\}$. The best subset $S$ is $\{e_{1,2}, e_{2, 3}\}$, i.e., the solid edges. The crossed edges $e_{2,3}$ and $e_{1,4}$ cannot be chosen for $S$ at the same time, according to the parallel constraint.}
	\label{fig:graph}
\end{figure}

$\widetilde{M_A}, \widetilde{M_B}, E$ make up a weighted bipartite graph $G = (\widetilde{M_A}\cup \widetilde{M_B}, E)$. 
Figure \ref{fig:graph} shows the graph $G$ of the example in Figure \ref{fig:example}.

\textbf{Optimization.} Using the notation introduced above, we restate the heuristic matching process as : \textit{Given a non-complete weighted  graph $G(\widetilde{M_A}\cup \widetilde{M_B}, E)$, find the instance set $S \subset E$ subject to (1) $|S|$ is maximized, and (2) $W(S)$, the total weight of $S$, is minimized.}

One cannot simply apply algorithms solving \textit{assignment problem} due to a \textit{parallel constraint}. Concretely speaking, \textit{for any two instances, if the antecedent-like pattern in one instance appears earlier than the other's, then its corresponding consequent-like pattern must also come earlier than the other's}. In the graph, this constraint requires no crossed edges in $S$, as is illustrated in Figure \ref{fig:graph}.

Suppose that the max $|S|$ is known as $s$ somehow, we can solve the following 0-1 integer programming problem:

\begin{subequations}
	\label{opt}
	\begin{align}
	\underset{x}{\text{minimize}}
	& \quad \sum{w_{i, j} x_{i, j}}   \label{opt:objective}\\
	\text{subject to} 
	& \quad \sum{x_{i, j}} = s \label{opt:const1}\\
	& \quad \sum_i{x_{i, j}} \le 1, \sum_j{x_{i, j}} \le 1 \label{opt:const2}\\
	& \quad  x_{i, j} + x_{k, l} \le 1, \forall\ i > k\ \text{and}\ l < j \label{opt:const3}\\
	& \quad x_{i, j} \in \{0, 1\} \label{opt:const4}
	\end{align}
\end{subequations}

The optimization variables $x_{i, j}$'s are $0,1$ variables, constrained by (\ref{opt:const4}), each of which represents the selection of corresponding instance $e_{i, j}$. (\ref{opt:const1}) restricts that $|S|$ is maximized as $s$. (\ref{opt:const2}) requires that at most one edge can be chosen in the graph $G$ with respect to the same vertex. (\ref{opt:const3}) refers to the parallel constraint.


Now consider how to solve for $s$. It is not the classical problem of \textit{maximum unweighted bipartite matching} due to the parallel constraint and therefore it cannot be easily solved by max/min flow algorithms. We formulate it as another optimization problem.

\begin{subequations}
	\label{opt2}
	\begin{align}
	\underset{x}{\text{maximize}}
	& \quad \sum{x_{i, j}}   \label{opt2:objective}\\
	\text{subject to} 
	& \quad \sum_i{x_{i, j}} \le 1, \sum_j{x_{i, j}} \le 1 \label{opt2:const1}\\
	& \quad  x_{i, j} + x_{k, l} \le 1, \forall\ i > k\ \text{and}\ l < j \label{opt2:const2}\\
	& \quad x_{i, j} \in \{0, 1\} \label{opt2:const3}
	\end{align}
\end{subequations}

The optimization problems are both 0-1 integer programming, which are NP-hard generally. Existing solvers (e.g. Matlab Optimization Toolbox) based on cutting plane method can handle these problems within a tolerable time. 


\subsection{Step 3. MDL-based Instance Scoring}

In this step, given the best instance set $S$, we first evaluate each instance by the similarity between it and the rule $r$ made up by $m_A$ and $m_B$, and then aggregate the results to the score of rule $r$ for further comparison. 

We first introduce the MDL-based scoring method, which is initially proposed in Shokoohi-Yekta et al. 2015.

Intuitively, the more similar the shape in the instance is with respect to the rule's consequent, the better the instance can support the rule. The Euclidean distance is the most widely accepted measure for similarity. However, the length of consequent varies in different rules, where Euclidean metric cannot fairly judge the differences between subsequences of different length.

Inspired by the Minimum Description Length principle that \textit{any regularity in a given set of data can be used to compress the data, i.e. to describe it using fewer symbols than needed to describe the data literally} \cite{grunwald2007minimum}, it is possible to take the rule's consequent as the regularity and measure how many bits can be saved by compressing the shape in the instances according to the regularity using Huffman coding. A concrete example can be found in Shokoohi-Yekta et al. 2015. 

To use MDL principle, the series must be digitized first, and let $dl(\cdot)$  be the coding length. The digitization loses little information of the raw data according. The number of bits saved for instance $e$ by encoding $\tilde{m_B}$ with respect to $r$'s consequent $m_B$ is as below:

\begin{equation}\label{mdl:1}
	bit\_saved(e, r) = dl(e.\tilde{m_B}) - dl(e.\tilde{m_B}|r.m_B)
\end{equation}

The above is the original version of MDL-based scoring method developed by Yekta et al.

We further take the ratio of antecedent-like pattern being matched into consideration. Intuitively, when the ratio is too small, indicating the number of matched instances is much less than the times that the antecedent is fired, the rule shouldn't be considered a good rule. Therefore, the final score for a rule $r$ is:

\begin{equation}\label{mdl:2}
	score(r) = \frac{s}{|\widetilde{M_A}|}\big(\sum_{e\in S}{bit\_saved(e, r)} - dl(r.m_B)\big)
\end{equation}

\section{Experiment Evaluation}
We evaluate our method on real open datasets. Top rules discovered by our method and the baseline method from the same training data are compared and analyzed. In addition, we also validate the applicability of our method on multiple series.

\subsection{Experiment Setup}
The baseline method (Y15) is the state-of-the-art work by Yekta et al. The experiment is conducted on two open metering datasets. One is Almanac of Minutely Power dataset (AMPds), mainly containing electricity stream data at one minute intervals per meter for 2 years of monitoring in a single house \cite{makonin2013ampds,makonin2016ampds}. 
 The other is UK Domestic Appliance-Level Electricity (UK-DALE) dataset, which records both the whole-house mains power demand every six seconds as well as power demand from individual appliances every six seconds from five houses \cite{UK-DALE}. 

\textbf{Settings.} Two groups of experiments are performed to (1) evaluate the prediction performance of our method and (2) validate the applicability of multiple series.

\begin{itemize}
	\item \textit{On Single Series.} We utilize the aggregate power series of \textit{clothes washer} and \textit{clothes dryer} for 1 year from AMPds, which is also used by Y15 as the experiment dataset. The series of first month is used to discover rules, while the rest is used for testing. We select $5$ top rules by each method and evaluate the prediction performance on the test data;
	\item \textit{One Multiple Series}. We attempt to discover rules from the separated power series of total $52$ appliances from the house 1 of UK-DALE dataset, such as boiler, laptop, etc. We run our method on each pair of the series to search for valid rules.
\end{itemize}

\textbf{Metric.} To measure the prediction performance of rules, we adopt the same metric $Q$ proposed by Yekta et al. as:
\begin{equation}
Q(r) = \frac{\sum_{i=1}^Nd(m_B, u_i)}{\sum_{i=1}^Nd(m_B, v_i)},
\end{equation}
where $N$ is the total firing number of the rule $r$ in the test data, and $d(m_B, x)$ is Euclidean distance between the consequent $m_B$ with the the shape beginning by position $x$. $u_i$ and $v_i$ are the $i$-th firing position and the $i$-th randomly chosen position, respectively. The denominator is used to normalize $Q$ to a value between $0$ and $1$. The final $Q$ is averaged after 1000 measurements.

A smaller $Q$ indicates a better prediction performance. The $Q$ close to 1 suggests the rule $r$ is no better than random guessing and $Q$ close to 0 indicates that the rule $r$ captures the structure in the data and predicts accurately.

\subsection{On Single Series}
To compare Y15 with our method, we select top 5 rules discovered by each method from the training data, and then evaluate them on the test data. 

\textbf{Result.} The top 5 ranked rules' $Q$s are listed in Table \ref{tab:q}. The top rules discovered by our methods are better than those by Y15. Specifically, our method outperforms Y15 by 23.9\% on average.
\begin{table}[b]
	\begin{tabular}{l c c c c c c}
		\toprule
		& 1 & 2 & 3 & 4 & 5 & Mean \\ 
		\midrule
		Y15 & 0.389 & 0.436 & 0.398 & 0.481 & 0.424 & 0.426  \\ 
		MBP & 0.340 & 0.299 & 0.337 & 0.310 & 0.341 & 0.324 \\ 
		\bottomrule
	\end{tabular} 
	\caption{The prediction performance $Q$ of top 5 rules on the test data.}
	\label{tab:q}
\end{table}

\begin{figure*}[t]
	
	\subfloat[The 5-th top rule $r_Y$ by Y15]{%
		\includegraphics[clip,width=\columnwidth]{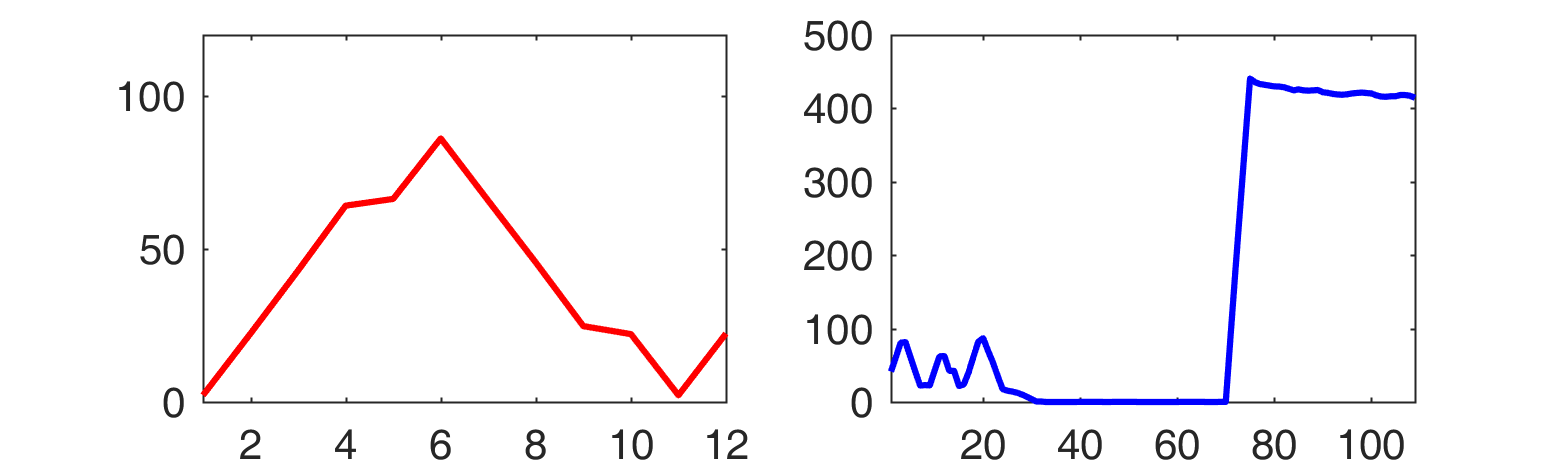}%
		\label{fig:ry}
	}
	\subfloat[The 5-th top rule $r_O$ by our method]{%
		\includegraphics[clip,width=\columnwidth]{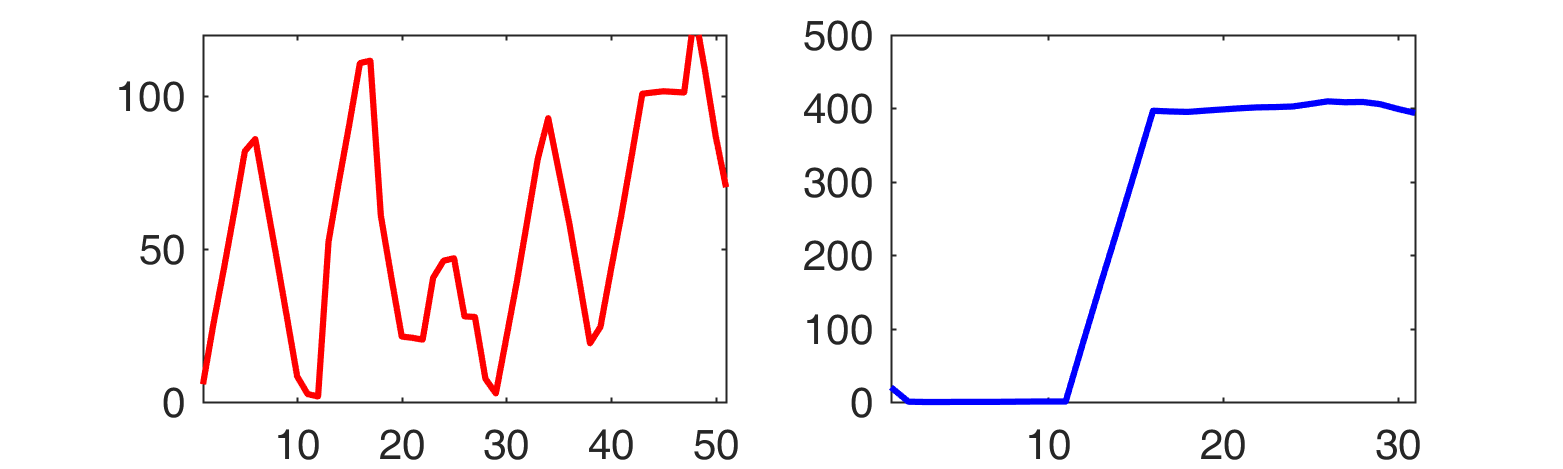}%
		\label{fig:ro}
	}
	\newline
	\subfloat[A firing by $r_Y$]{%
		\includegraphics[clip,width=\columnwidth]{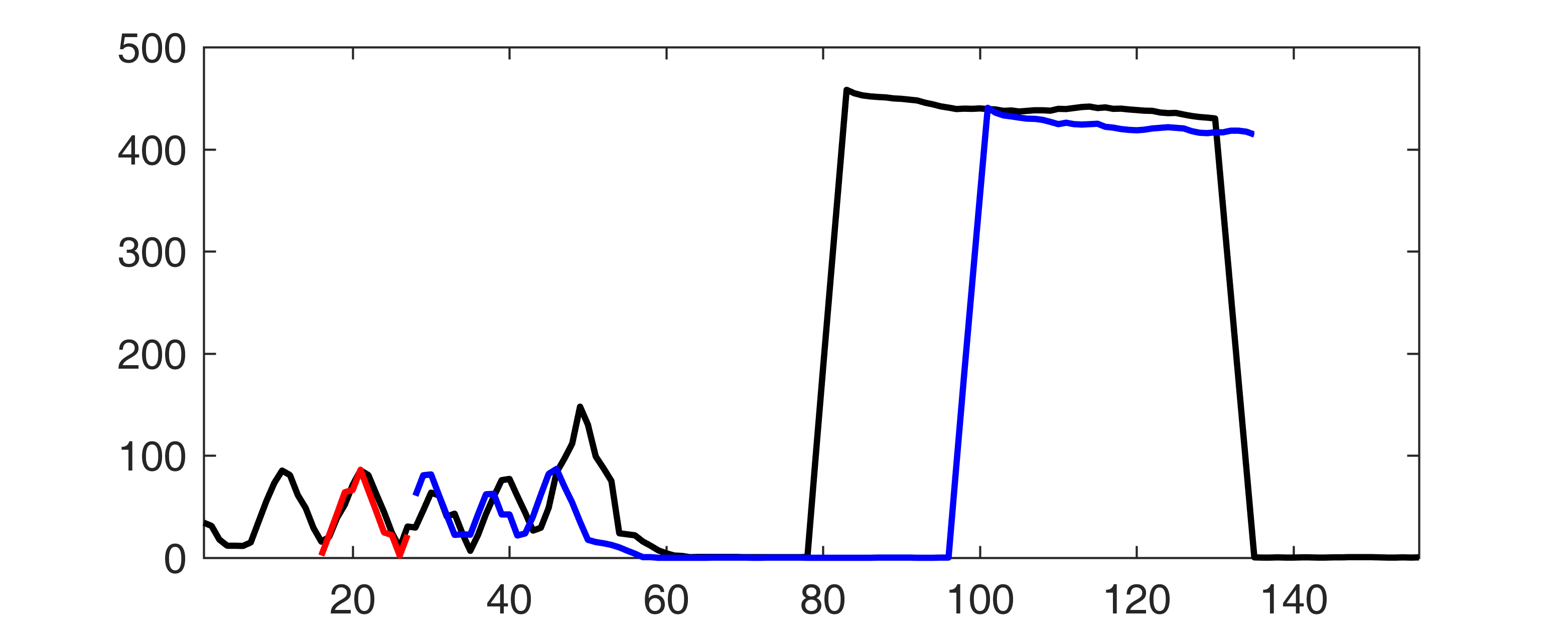}%
		\label{fig:fire_ry}
	}
	\subfloat[A firing by $r_O$]{%
		\includegraphics[clip,width=\columnwidth]{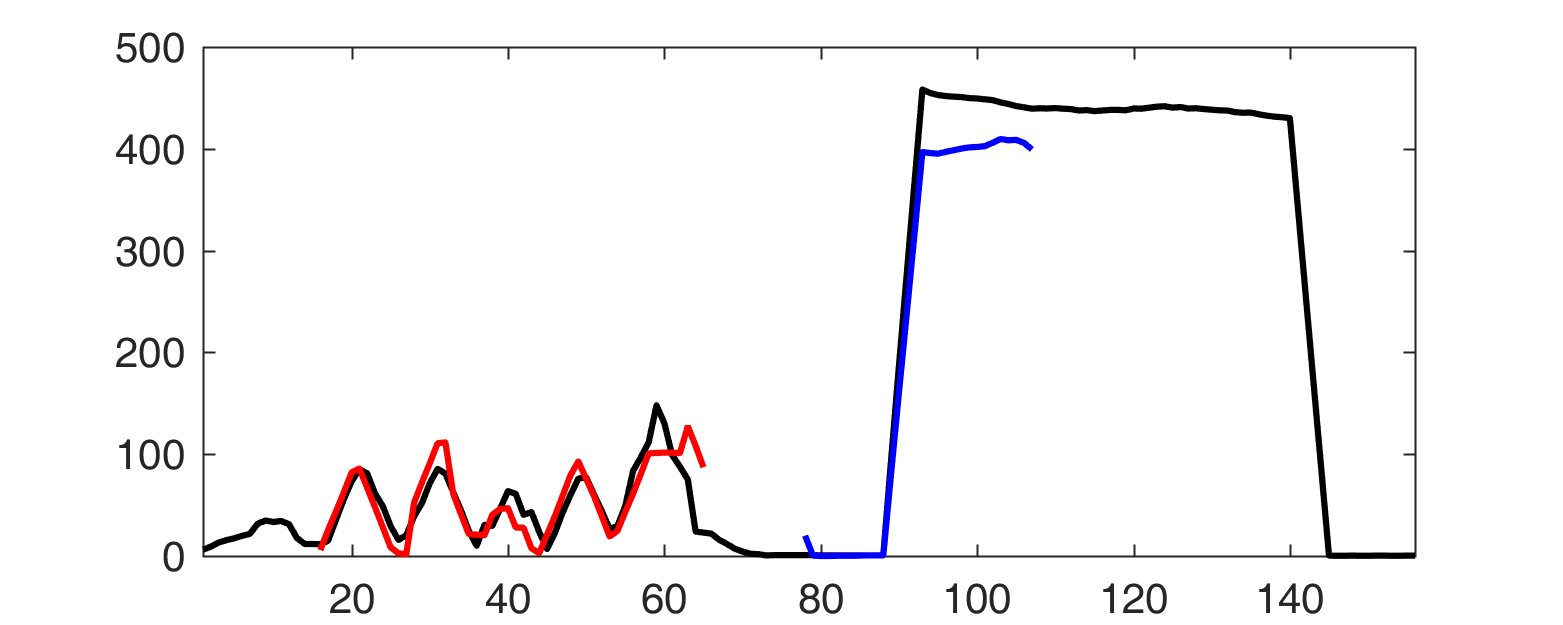}%
		\label{fig:fire_ro}
	}
	\caption{The two 5-th top rules with the prediction result around a same position. In \subref{fig:ry} and \subref{fig:ro}, the red patterns are the antecedent of the rule, while the blue ones are the consequent. The rule's threshold $\theta$ is set to 5 and max length time interval $\tau$ is set to 300 in both methods. \subref{fig:fire_ry} \subref{fig:fire_ro} depict the prediction results around the same position by both rules. The black curve is the real time series. The red curve shows the position where the rule is fired, while the blue curve is the best prediction during the max time interval.}
	\label{fig:top}
\end{figure*}

\textbf{Comparison.} To demonstrate the reason why the rules discovered by our method outperform those by Y15, we take a close look at the 5-th top rule, $r_Y$ (by Y15) and $r_O$ (by this paper), and scrutinize their prediction results on the test data in Figure \ref{fig:top}.

As is shown in Figure \ref{fig:ry} and \ref{fig:ro}, the 5-th top rules $r_Y$ and $r_O$ are quite different from each other though they are describing the same thing\footnote{Actually, they both imply the fact that the clothes dryer is often used after the clothes washer.}. $r_Y$ comes from a splitting point at $10\%$-th of a $120$-long subsequence, whose antecedent is $12$ in length and consequent is $108$ in length, whereas $r_O$ takes two motifs as its antecedent and consequent, whose lengths are 50 and 30 respectively. In contrast, $r_Y$ has more reasonable antecedent and consequent. To illustrate, consider the case in Figure \ref{fig:fire_ry} and \ref{fig:fire_ro}. 

The antecedents of both rules present a good match and trigger both rules around the same position. However, $r_O$ gives an accurate prediction,
 while $r_Y$ predicts a shape with a clear discrepancy to the real series. The interval before $r_Y$'s consequent is so long that the consequent misses the best matching position. Intuitively,  $r_Y$'s consequent can be viewed as $r_O$'s consequent appended by a piece of noise series, which results in the mismatch of the consequent.

The inaccurate prediction has its root in the splitting method of Y15, which  inevitably adds some extra noises to the ``real'' antecedent/consequent, because the splitting method cannot position the boundaries of the antecedent/consequent. Our method, however, directly finds the key part of the series, i.e. motif, as the rule's antecedent/consequent. 

Additionally, any rule discovered by Y15 is a split of a motif in this experiment. Since the split parts are also frequent patterns in the series, they can be discovered as motifs, i.e. the elements of $M_A$ and $M_B$. Therefore, rules discovered by Y15 can also be found by our method.

\textbf{Discussion.} 
 In the electricity datasets, zero series\footnote{The values in the series are almost all $0$s, indicating no appliance is being used.} is recognized as motif by MK algorithm because it is also a frequent pattern (though meaningless). To avoid such motifs, we sort the discovered motifs by the ``roughness'' of the shape and choose the top ones. Commonly, the sorting process, mentioned in the line 2 of Algorithm 1, is relevant with the characteristics of the time series.

\subsection{On Multiple Series}

We attempt to discover rules in multiple appliance series from the UK-DALE datasets, which include $52$ kinds of appliances. The original data is sampled every six seconds, and we resample it per minute for efficiency consideration.

\begin{figure}[t]
	\includegraphics[clip,width=\columnwidth]{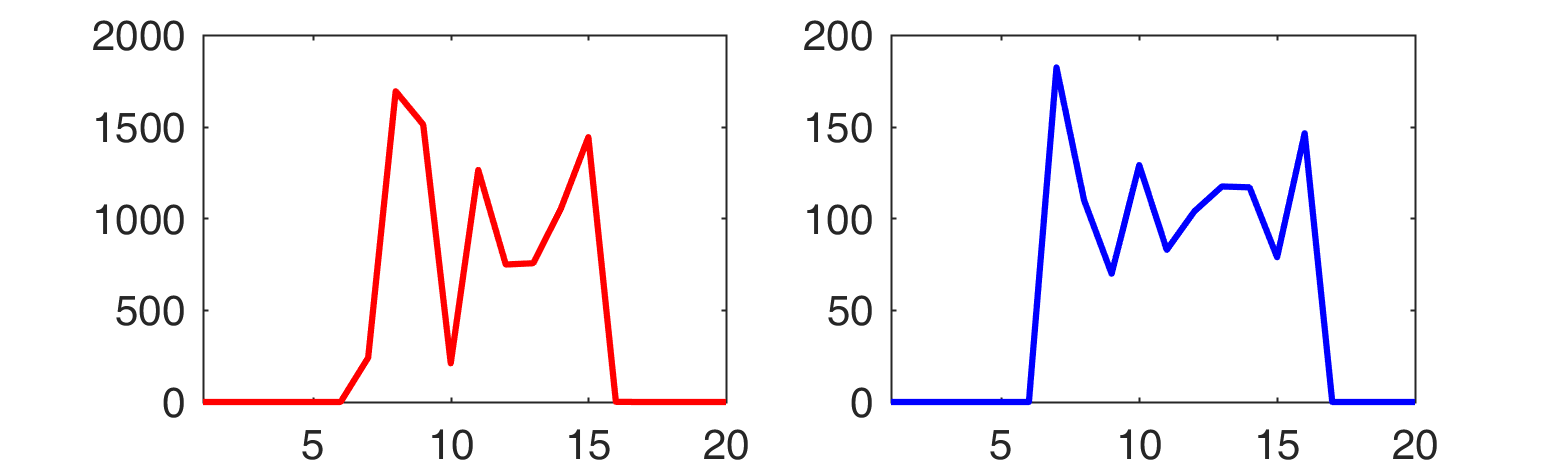}%
	\caption{A rule $r_E$ discovered in the series of hair dryer and straightener. The red curve is the antecedent, while the blue curve is the consequent.}
	\label{fig:rule}
\end{figure}

\textbf{Result.} A serviceable rule $r_E$ discovered is from the power series of hair dryer and straighteners, the antecedent and consequent of which are  shown in Figure \ref{fig:rule}.

\begin{figure}[t]
	\includegraphics[clip,width=\columnwidth]{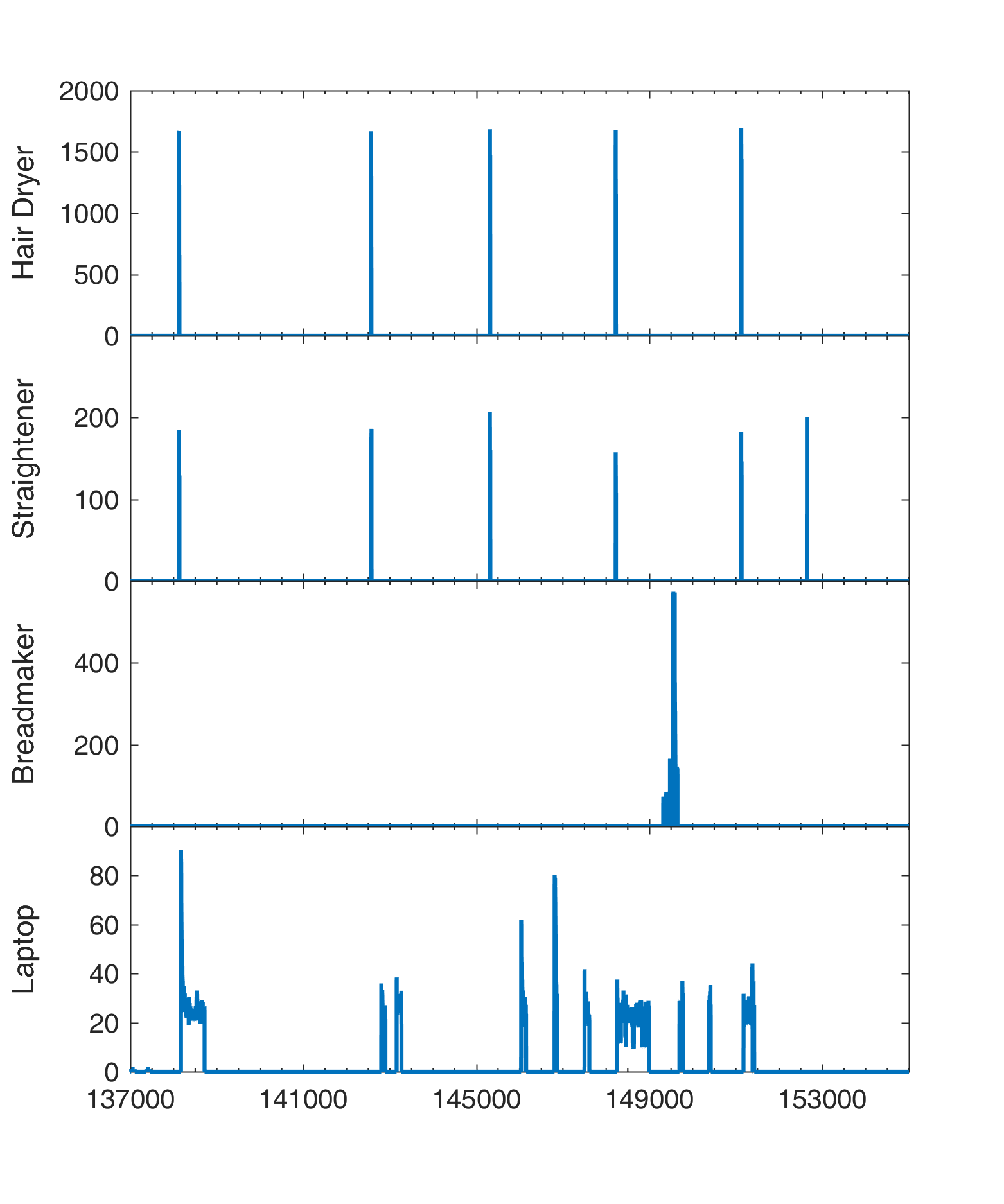}%
	\caption{The overviews of four appliances' power series.}
	\label{fig:series}
\end{figure}

The rule $r_E$ describes the relationship between the usage of hair dryer and straightener. 
 An interesting fact is that the rule's antecedents and consequents are interchangeable, coinciding the common sense that the two appliances, hair dryer and straightener, are often used at the same time. To illustrate  $r_E$ concretely, we list an overview of 4 power series in Figure \ref{fig:series}, including hair dryer, straightener, breadmaker and laptop. The series of hair dryer and straightener are well matched, whereas the rest combinations are ranked relatively lower.


\section{Conclusion and Discussion}
In this paper, we have introduced a novel rule-based prediction method for real-valued time series from the perspective of motifs. We preliminarily explore the possibility of relating two motifs as rule candidate.  It first leverages motif discovery to segment the time series precisely for  seeking recurring patterns as antecedents/consequents, and then investigates the underlying temporal relationships between motifs by combing motifs as rule candidates and ranking them based on the similarities. 

However, as is mentioned before, this work itself is incomplete and will be refined. We further consider the following two problems:

First, current experiment mainly uses one kind of open dataset, i.e. household electricity usage. We will search for more open datasets to comprehensively evaluate the performance of our method.

Second, in this work we evaluate each rule from the perspective of prediction. However, prediction is only a single aspect of a rule. We will try to develop more metrics that can reveal the inner connections within rules.


\section{Acknowlegements}
This work is supported by the program JS71-16-005.

\bibliography{he447.bib}
\bibliographystyle{aaai}
\end{document}